\documentclass[11pt,a4paper]{article}
\usepackage[hyperref]{naaclhlt2018}
\usepackage{times}
\usepackage{latexsym}
\usepackage{amsmath}
\usepackage{booktabs}
\usepackage{graphicx}
\usepackage{subfig}

\usepackage{url}

\aclfinalcopy 
\def\aclpaperid{3}

\title{Can Entropy Explain Successor Surprisal Effects in Reading?}

\author{Marten van Schijndel\\
  Department of Cognitive Science\\
  Johns Hopkins University\\
  {\tt vansky@jhu.edu} \\\And
  Tal Linzen\\
  Department of Cognitive Science\\
  Johns Hopkins University\\
  {\tt tal.linzen@jhu.edu}}

\date{}

\begin{document}
\maketitle
\begin{abstract}
    Human reading behavior is sensitive to surprisal: more predictable words tend to be read faster. Unexpectedly, this applies not only to the surprisal of the word that is currently being read, but also to the surprisal of upcoming (successor) words that have not been fixated yet. This finding has been interpreted as evidence that readers can extract lexical information parafoveally. Calling this interpretation into question, \newcite{angeleetal15} showed that successor effects appear even in contexts in which those successor words are not yet visible. They hypothesized that successor surprisal predicts reading time because it approximates the reader's uncertainty about upcoming words. We test this hypothesis on a reading time corpus using an LSTM language model, and find that successor surprisal and entropy are independent predictors of reading time. This independence suggests that entropy alone is unlikely to be the full explanation for successor surprisal effects.
\end{abstract}

\section{Introduction}

One of the most robust findings in the reading literature is that more predictable words are read faster than less predictable words \cite{ehrlich1981contextual}. Word predictability effects fit into a picture of human cognition in which humans constantly make predictions about upcoming events and test those predictions against their perceptual input \cite{bar2007proactive}.

While the effect of the predictability of the current word ($w_t$) on the reading time at $w_t$ is not controversial, there is a spirited debate in the eye movement literature as to whether reading time at $w_t$ is affected by the predictability of the \textit{successor} word, $w_{t+1}$ \cite{drieghe11}. Reading is characterized by a series of fixations, which bring a single word into the center of the visual field (the fovea), where visual acuity is highest. Effects of successor predictability have been taken to indicate that readers are able to process words parafoveally, that is, even when those words are not fixated \cite{kliegletal06}. Such an empirical finding would appear to constitute evidence against serial attention shift models such as E-Z Reader \cite{reichle2003ez}, in which attention is directed at a single word at a time, and in favor of models such as SWIFT \cite{engbert2002dynamical}, in which attention can be distributed over multiple words at the same time.

This interpretation of successor predictability effects was called into question by \newcite{angeleetal15}, who showed that the predictability of word $w_{t+1}$ affected reading time at $w_{t}$ even when $w_{t+1}$ was masked and was not visible until the reader fixated on it directly. A similar result was found by \newcite{vanschijndelschuler17} in self-paced reading, a paradigm which similarly precludes parafoveal preview. Short of ascribing psychic abilities to readers, then, the only possible explanation for these findings is that what appears to be an effect of the predictablity of $w_{t+1}$ is a confound driven by the relationship between the predictability of $w_{t+1}$ and an underlying property of $w_t$.

\newcite{angeleetal15} hypothesized that the property of $w_t$ that is confounded with the predictability of $w_{t+1}$ is the reader's \textbf{uncertainty} about the words that could follow $w_t$, but they did not test this hypothesis. The present paper directly evaluates the relation between successor surprisal and uncertainty estimated from a single RNN language model \cite{gulordavaetal18}. We use a self-paced reading corpus \cite{futrelletal18}, in which parafoveal preview is unavailable. To anticipate our results, we do not find evidence that the effect of successor surprisal can be reduced to uncertainty. We then explore the hypothesis that processing limitations, which lead to uncertainty being calculated over a restricted number of probable words rather than over the entire vocabulary, could account for these conflicting results, with similarly negative results. We conclude that uncertainty is unlikely to be the only explanation for successor surprisal effects. 

\section{Surprisal and entropy}

The relationship between the reading time at word $w_t$ and the conditional probability of $w_t$ is logarithmic \cite{smith2013effect}; in other words, if we use \emph{surprisal} \cite{hale01} as our probability measure:
\begin{equation}
\text{surprisal}(w_t)=-\text{log P}(w_t\mid w_{1\ldots t-1})\label{eqn:surp}
\end{equation}

\noindent then there is a linear correlation between RT($w_t$) and surprisal($w_t$).
Surprisal has been shown to be a strong predictor of reading time in linear regression models \cite[e.g.,][]{dembergkeller08,roarketal09}.

Successor surprisal is simply the surprisal of the next observation in a sequence:
\begin{align}
  \text{succ.\ surprisal}(w_t)&=-\text{log P}(w_{t+1}\mid w_{1...t})\label{eqn:succsurp}\\
        &=\text{surprisal}(w_{t+1})
\end{align}

\noindent Finally, the entropy at $w_t$ is defined as follows:
\begin{align}
&H(w_t) = E[\text{surprisal}(w_{t+1})]\label{eqn:esurp}\\
  &= -\!\!\!\!\!\!\sum_{w_{t+1}\in V}\!\!\!\text{P}(w_{t+1}\!\mid\! w_{1...t})\ \text{log P}(w_{t+1}\!\mid\! w_{1...t})\label{eqn:fullH}
\end{align}

As mentioned in the introduction, \citet{angeleetal15} hypothesized that the entropy at $w_t$ is the underlying cause for successor ($w_{t+1}$) surprisal effects on $w_t$. This is a plausible hypothesis: the expected successor surprisal in a given context is the entropy at $w_t$ (Equation~\ref{eqn:esurp}), so in the limit, successor surprisal should be the same as the entropy over possible continuations when averaged over a corpus.
In this hypothetical limit-case, we would directly observe Equation~\ref{eqn:fullH} in the data, as the sequence $w_{1...t+1}$ occurred exactly the expected number of times in the corpus.
In practice, with a finite set of observations $T$ which are regressed simultaneously, successor surprisal provides a Monte Carlo estimator of entropy in that corpus:
\begin{align}
  \hat H(T)&\approx-\sum^{\mid T\mid}_{t=1}\frac{1}{\mid\!\! T\!\!\mid}\ \text{log P}(w_{t+1}\mid w_{1...t})\\
  &=\sum^{\mid T\mid}_{t=1}\frac{1}{\mid\!\! T\!\!\mid}\ \text{surprisal}(w_{t+1})\label{eqn:succsurpapprox}
\end{align}
Therefore, if uncertainty over possible continuations influences reading time, then successor surprisal could be correlated with reading time simply due to its relationship with corpus-level entropy.

Importantly for the present study, if the relationship to uncertainty is the sole underlying reason that successor surprisal can predict reading time, successor surprisal should be a worse predictor of reading time than entropy when the same model distribution $q$ is used to compute both measures.
This claim follows directly from Equation~\ref{eqn:succsurpapprox}: if entropy $H_q$ is the true generator of the data, then it should always be a better predictor than some corpus-level approximation $\hat H_q$ due to noise from the Monte Carlo process.

The syntactic language models used in previous reading studies could not compute successor surprisal and entropy from the same conditional probability distribution, precluding a direct test of this hypothesis; in particular, while the \citet{roarketal09} parser can compute both surprisal and entropy, it estimates them using two different probability distributions due to its use of beam search.

\begin{figure}
\includegraphics[scale=0.4]{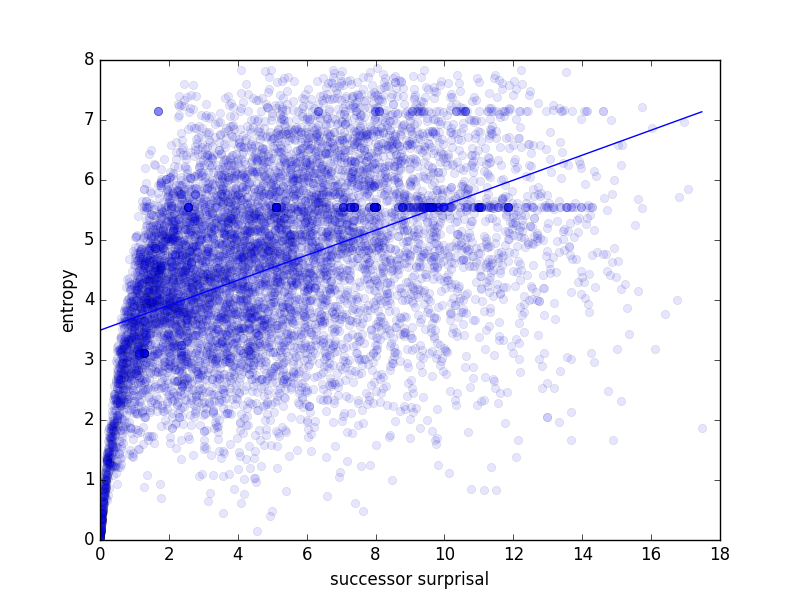}
\caption{Successor surprisal plotted against entropy for each word in the Natural Stories Corpus.
The Pearson correlation is 0.45, providing empirical validation of the theoretically strong limit-case relation between entropy and successor surprisal.}\label{fig:basecorr}
\end{figure}

\section{Method}\label{sec:method}

\paragraph{Language model:} In contrast with previous work, which used grammar-based language models, we used a single recurrent neural network (RNN) language model to compute entropy and successor surprisal from the same conditional probability distribution.  The language model we used was trained by \citet{gulordavaetal18} on 90 million words from the English Wikipedia.
The model had two LSTM layers with 650 hidden units each, 650-dimensional word embeddings, a dropout rate of 0.2 and batch size 128, and was trained for 40 epochs (with early stopping).

Unlike grammar-based language models, RNN language models do not explicitly construct syntactic dependencies, which are essential in human sentence comprehension. However, recent work has shown that RNN language models are nevertheless sensitive to the probability of syntactic structures \cite{linzenetal16,vanschijndellinzen18,wilcox2018rnn}, tentatively suggesting that they are an adequate substitute for modeling human reading behavior. Importantly, they have the added benefit that all our measures of interest are easy to calculate using Equations~\ref{eqn:surp}, \ref{eqn:succsurp}, and \ref{eqn:fullH} on the model's softmax layer, which provides a conditional probability distribution over the upcoming word given the preceding words.

\paragraph{Data:} The test domain in this work is the Natural Stories Corpus \cite{futrelletal18}.
The  corpus is a set of 10 texts (485 sentences) written to sound fluent while still containing many low-frequency and marked syntactic constructions.
The sentences within each text were presented in order, and self-paced reading data were collected from 181 native English speakers.
We used one third of the sentences for exploration while two thirds were set aside for statistical confirmation.
We omit any words consisting of multiple tokens (e.g., \emph{do$\cdot$n't} and \emph{boar$\cdot$!$\cdot$'}).
 In this paper, all statistical testing was done on the held-out partition.

 \section{Results}\label{sec:results}

 \paragraph{Successor surprisal is moderately correlated with entropy:} We first tested the degree to which the Monte Carlo estimation produces a correlation between entropy and successor surprisal when each is computed with the same probability model (i.e.\ the LSTM language model described in Section~\ref{sec:method}) and found that the measures were moderately correlated, with Pearson's $r = 0.454$ (see Figure~\ref{fig:basecorr}).
This moderate correlation between the two measures could plausibly explain the successor surprisal effects on reading time that have been observed in previous studies. 

\paragraph{Successor surprisal predicts reading time:} Before testing whether entropy can account for the effectiveness of successor surprisal in predicting reading time, we first verified that our successor surprisal measure was positively correlated with reading time as observed with the language models used in previous work \cite{angeleetal15,vanschijndelschuler16,vanschijndelschuler17}.

Following previous studies, we used a linear mixed effects regression approach. Unlike linear regression, in which the error term is assumed to come from a single normal distribution, this approach takes into consideration clustered errors that are due to the variability across the particular participants and words in the sample (``random effects''). This makes it possible to estimate the effect of theoretically relevant ``fixed effects'' in a way that is more likely to generalize to new items and participants. We used the \emph{lme4} R package \cite{r-lme4} to perform the regression, and included fixed effects for word length, sentence position, unigram frequency, surprisal, and successor surprisal.
Unigram frequencies were estimated from the Gigaword corpus \cite{graffcieri03}.
We included random intercepts for each word and subject, and by-subject random slopes for each fixed effect.%
\footnote{We also ran all of the analyses reported in this paper on the exploratory partition without the random word intercept and obtained qualitatively similar results.}
All predictors were z-transformed before fitting the models.
We compared the log-likelihood of the data under that model to the log-likelihood of one without the fixed effect for successor surprisal to determine the significance of successor surprisal as a fixed effect predictor of reading time.

Successor surprisal was significant as a predictor ($\hat\beta=4.3$, $\hat\sigma=0.52$, $\chi^2(1) = 58$, $p < 0.001$), suggesting that the previously observed relationship between successor surprisal and reading time holds when successor surprisal is computed with our LSTM language model.
We note that, in this self-paced reading setting, the regression coefficient of successor surprisal was quite large: it was over half that of the coefficient of $w_t$ surprisal ($\hat\beta=6.0$) and rivaled that of unigram frequency ($\hat\beta = 5.1$).

 \begin{table}[t]
   \begin{tabular}[t]{l r r r}
          \toprule
    & $\hat\beta$ & $\hat\sigma$ & $t$\\
          \midrule
    (Intercept)            & 331.66  &   6.31 & 52.56\\
    Sentence position      &   0.72  &   0.51 & 1.41\\
    Word length            &   4.74  &   1.00 & 4.73\\
    Surprisal              &   5.67  &   0.57 & 9.88\\
    Unigram frequency      &   4.94  &   1.18 & 4.17\\
    Successor surprisal    &   3.26  &   0.39 & 8.34\\
    Entropy                &   3.12  &   0.55 & 5.68\\
        \bottomrule
       \end{tabular}
   \caption{Fixed effect coefficients from fitting self-paced reading times. Since predictors were z-transformed, the $\hat\beta$ coefficients indicate the change in ms per standard deviation of each predictor.}\label{tab:futsurp_and_entropy}
   \end{table}

\paragraph{Entropy and successor surprisal account for different portions of the variance:}

If successor surprisal is only predictive of reading time because it approximates entropy as hypothesized by \citet{angeleetal15}, then entropy should not only be predictive of reading time, but it should also obviate successor surprisal as a predictor since the approximation (successor surprisal) would only get credit for indirectly modeling part of the influence of entropy.
To test this, we added entropy as a fixed effect and as a by-subject random slope to our linear-mixed effects model.
Comparing the fit of that model to the fit of a model without each fixed effect of interest, we found that successor surprisal and entropy were both significant predictors of reading time (both $p < 0.001$; see Table~\ref{tab:futsurp_and_entropy}); thus the hypothesis that the effect of entropy should subsume the effect of successor surprisal was not borne out.

\section{Bounded entropy}

Entropy and successor surprisal both accounted for independent portions of the variance in reading time in Section \ref{sec:results}. Could they both provide indirect approximations of underlying reader uncertainty?
So far we computed entropy over the complete distribution of possible upcoming words (\emph{total entropy}). In this section, we explore the possibility that processing limitations cause readers to consider only the best $K$ continuations in the psychological process that causes uncertainty effects \citep[see bounded rationality,][]{simon82,jurafsky96}.
If this is the case, then total entropy and its successor surprisal approximation could both be predictive of reading time because of their joint correlation with the bounded entropy computed by humans. 

\begin{table}
\begin{center}
  \begin{tabular}{l r r}
      \toprule
    $K$ & Successor surprisal & Total entropy\\
      \midrule
  5 & 0.212 & 0.541\\
  50 & 0.335 & 0.820\\
  500 & 0.397 & 0.947\\
  5000 & 0.434 & 0.992\\
  50000 & 0.454 & 1\\
      \bottomrule
  \end{tabular}
  \caption{Correlations between (Center) best-$K$ entropy and successor surprisal and (Right) best-$K$ entropy and total entropy when best-$K$ entropy is computed over the most probable $K$ continuations.}\label{tab:topkcorr}
  \end{center}
\end{table}

To test this hypothesis, we computed entropy over just the best 5, 50, 500, and 5000 continuations in every context. The full vocabulary size of the model was 50000 (plus an UNK token). Successor surprisal was always computed over the full vocabulary so that every observation could be assigned a successor surprisal value.

Entropy was most correlated with successor surprisal when both measures were computed over the entire vocabulary (Table~\ref{tab:topkcorr}).
This is a plausible finding given Equation~\ref{eqn:succsurpapprox}, which indicates that successor surprisal provides a Monte Carlo approximator of the entropy of that same distribution (recall that successor surprisal was calculated over the full vocabulary).
It may still be the case, however, that reading time is best predicted by one of the bounded entropy measures. For example, best-50 entropy still has a moderate correlation to successor surprisal (0.335) and a strong correlation to total entropy (0.82); it is possible that total entropy and successor surprisal both predicted reading time thanks to an underlying joint correlation with best-50 entropy.

To test whether that is the case, we used our bounded entropy variants to predict reading time, following the procedure of Section \ref{sec:results}.%
\footnote{For these analyses, we omit by-subject random slopes for sentence position, surprisal, and unigram frequency in order to ensure that all 5 models converge. Leaving all random slopes in the models produces similar qualitative results in those models that do converge.}
Bounded entropy was a consistently poorer predictor of reading time than total entropy (see Table~\ref{tab:topkspr}).
This suggests that humans may be sensitive to uncertainty over a large number of possible continuations.
Moreover, successor surprisal improved as a predictor of reading time as $K$ decreased and the predictive value of bounded entropy weakened. This trade-off indicates that some of the variance in reading time is explained by both measures, which suggests that the predictivity of successor surprisal in previous studies was at least partially driven by reader uncertainty \citep[in line with][]{angeleetal15}.
However, the continued predictivity of successor surprisal in the presence of entropy indicates that there are likely other factors involved as well.
For example, it may be that readers make predictions of varying granularity depending on context or attention level. 
That is, in cases where readers make a prediction based on the best $K$ continuations and $K$ is similar to the bound for computing entropy, then entropy may help predict reading time.
Successor surprisal could help absorb variance due to a mismatch between reader $K$ and the model's $K$.

\begin{table}
    \centering
  \begin{tabular}{l r r r r}
    \toprule
    $K$ & $\hat\beta_{H}$ & $\hat\sigma_{H}$ & $\hat\beta_{s}$ & $\hat\sigma_{s}$\\
    \midrule
5     &    3.10  &   0.70 & 3.89 & 0.53\\
50    &    3.27  &   0.71 & 3.81 & 0.54\\
500   &    3.89  &   0.70 & 3.65 & 0.54\\
5000  &    4.39  &   0.70 & 3.53 & 0.54\\
50000 &    4.57  &   0.70 & 3.48 & 0.54\\
    \bottomrule
  \end{tabular}
  \caption{Entropy ($H$) and successor surprisal~($s$) coefficients in the Section~\ref{sec:results} RT regression model for the exploratory data partition, when $H$ is calculated over the $K$ most probable continuations.} 
  \label{tab:topkspr}
\end{table}

\section{Related work}

Previously, \citet{vanschijndelschuler17} performed a similar analysis to the reading time analysis in Section~\ref{sec:results} in this paper using probabilistic context-free language models.
They were forced to compute entropy and successor surprisal with separate models because entropy computation using a grammar-based model requires estimation of uncertainty over both words and parsing actions, and is therefore very computationally expensive.
While they also found that entropy and successor surprisal independently predicted reading time, their use of multiple language models means that the independent predictivity in their study could arise from differences in their underlying models instead of from multiple independent reading time influences.
In contrast, we wanted to directly compare the measures as estimated by a single model to provide a stronger test of the original hypothesis of \citet{angeleetal15}.

\citet{frank13} conducted a related reading time analysis which studied the relationship between entropy reduction \cite{hale06} and surprisal as computed by neural network language models.
Entropy reduction is a measure of how uncertainty about the future changes after an observation compared to before that observation.
Since entropy reduction involves the difference between two levels of uncertainty, it is a distinct measure from the amount of uncertainty (entropy) over upcoming observations which we studied in this paper.
That is, the fact that uncertainty is reduced after an observation says nothing about the total amount of uncertainty experienced by a reader after that lessening takes place.%
\footnote{For example, $H(w_t) - 2 = H(w_{t+1})$ does not convey how large $H(w_t)$ or $H(w_{t+1})$ are. This amount of entropy reduction (2) could equally occur in a context of high uncertainty or in one of low uncertainty.}
\citet{frank13} found that entropy reduction and surprisal are also distinct measures with independent reading time predictivity, similar to the findings of entropy and successor surprisal in the present paper.

\citet{frank13} also tested how the relationship between entropy reduction and surprisal changed when the uncertainty used to estimate entropy reduction was computed over more than just the single next upcoming observation; he found that the predictive value of entropy reduction improves when entropy is computed over multiple future words. 
However, in the context of the present paper, \citet{angeleetal15} observed a direct relationship between the predictability of a single word ($w_{t+1}$) on the reading time of the preceding word ($w_t$).
Further, \citet{vanschijndelschuler16} previously found that successor surprisal best predicts reading time when computed over just the upcoming one or two words even when parafoveal preview is possible, so it seems unlikely that computing entropy over longer upcoming sequences like \citet{frank13} could explain the remaining successor surprisal influence on self-paced reading observed in this study.
Therefore, since the goal of the present paper was to test the \citet{angeleetal15} hypothesis that the entropy over $w_{t+1}$ could be the driving influence behind successor surprisal, we focused on testing the relationship between the reading time at $w_t$ and measures of entropy over $w_{t+1}$ and did not explore the influence of uncertainty over words beyond $w_{t+1}$.

\section{Discussion}

This paper has used surprisal and entropy estimates from a neural network language model to test the hypothesis that successor surprisal effects in reading can be reduced to reader uncertainty. Successor surprisal and uncertainty accounted for partly non-overlapping portions of the variance in reading time. 
We interpret our finding of non-overlapping influences as a strong indictation that the predictivity of successor surprisal is not solely driven by uncertainty over the next word.

However, the portions of variance captured by entropy and successor surprisal are not completely disjoint: replacing entropy with bounded variants based on the best $K$ continuations led to weaker predictive power for entropy and a stronger relationship between successor surprisal and reading time, lending support to the hypothesis that entropy is at least a contributing factor in the predictivity of successor surprisal. Finally, the finding that uncertainty was a better predictor of reading time when it was computed over the entire vocabulary rather than just the best $K$ continuations suggests that readers may make a large number of continuation predictions simultaneously.

\bibliography{bibliography}
\bibliographystyle{acl_natbib}
\end{document}